\pdfoutput=1
\documentclass[]{article}
\usepackage[margin=1in]{geometry}
\usepackage{amsmath}
\usepackage[hidelinks]{hyperref}
\usepackage{graphicx}
\usepackage{multirow}
\usepackage{tabularx}
\usepackage{booktabs}
\usepackage{url}
\usepackage{amssymb}
\usepackage{float}
\usepackage{fixme}
\usepackage{cleveref}
\usepackage{listings}
\fxsetup{
    status=draft,
    layout=inline,
    theme=color
}
\usepackage{CJKutf8}
\usepackage[backend=biber,sorting=none]{biblatex}

\crefname{figure}{Figure}{figures}
\Crefname{figure}{Figure}{Figures}
\crefname{section}{Section}{sections}

\newcommand{\resultgridfiguresscale}{.92}

\renewenvironment{abstract}
 {\small
  \begin{center}
  \bfseries \abstractname\vspace{-.5em}\vspace{0pt}
  \end{center}
  \list{}{
    \setlength{\leftmargin}{5mm}
    \setlength{\rightmargin}{\leftmargin}
  }
  \item\relax}
 {\endlist}

\title{Mixture of Tunable Experts - Behavior Modification of DeepSeek-R1 at Inference Time}

\usepackage{authblk}
\author[1,*]{Robert~Dahlke}
\author[1,*]{Henrik~Klagges}
\author[1]{Dan~Zecha}
\author[1]{Benjamin~Merkel}
\author[1]{Sven~Rohr}
\author[1]{Fabian~Klemm}

\affil[1]{TNG Technology Consulting GmbH, Germany}
\affil[*]{joint first authorship\linebreak\linebreak \href{mailto:research@tngtech.com}{\nolinkurl{research@tngtech.com}}}

\date{}

\newcommand{\classemph}[1]{\texttt{#1}}
\newcommand{\promptquote}[1]{\textit{#1}}

\newcommand{\fooendpromptplaceholder}{\}}
\newcommand{\beginpromptplaceholder}{\{}

\begin{document}

\maketitle

\begin{abstract}
We present the Mixture-of-Tunable-Experts (MoTE), a method that extends the Mixture-of-Experts architecture of Large Language Models (LLMs). Without additional training, MoTE enables meaningful and focused behavior changes in LLMs on-the-fly during inference time.

By analyzing the digital LLM brain of DeepSeek-R1 using a technique we dub ``functional Token Resonance Imaging" (fTRI) – inspired by fMRI and using prompts designed to elicit specific behavior (e.g., \promptquote{``What happened \beginpromptplaceholder time\fooendpromptplaceholder  \beginpromptplaceholder place\fooendpromptplaceholder ?"}) – we empirically identify distinctive experts associated with behaviors like refusal responses.

Using MoTE we are able to intervene and control such specific behavior.
We switched off the top 10 most refusal-relevant experts (0.07\% of R1's 14,848 routed experts), achieving a 52\% refusal reduction on sensitive reference prompts without performance degradation on MT-Bench.
Random expert deactivation resulted in smaller behavioral shifts with increased noise, whereas forced expert activation led to significantly higher refusal rates.
With MoTE we were also able to successfully switch the model's chain-of-thought reasoning language from English to Chinese in 10\% of our test prompts.

Our approach shares similarities with sparse autoencoders (SAEs) in terms of explainability and steerability.
Unlike SAEs, MoTE does not require large training efforts, as within MoEs with a vast number of experts, specialization already emerged naturally during pretraining.

Our findings suggest that significant functional mechanisms in Mixture-of-Experts architectures can at least partially be localized in a small number of specific experts, rather than being distributed throughout the model's weights.
Expert subgroups can be tuned to trigger significant behavior variations, providing insights into the inner workings of LLMs.
\end{abstract}

\section{Introduction}

Large Language Models are rapidly becoming indispensable tools for millions of companies and organizations, hundreds of millions of people, and entire nation states.
Widely differing needs create demand for adapting LLMs in behaviors and capabilities beyond mere model selection.
There are several known classes of such adaptability:

\begin{enumerate}
    \item \emph{Prompt engineering} works at the regular chat interface of LLMs and thus can be applied to almost any LLM, even for closed-source systems where weights are not available. On the provider side, creating system prompts gives a lot of delivery flexibility. On the user side, prompting the LLM creatively is the default way to coax additional behaviors out of it. With carefully constructed prompts, prompt engineering can be taken to the extremes of ”jail breaking”, which is almost an art form perfected by disguised experts such as the famous Pliny the Prompter.
    \item \emph{Goal-oriented finetuning} is significantly harder to implement and requires a higher degree of sophistication. Dataset curation is crucial, along with either direct access to the base model weights or access to a dedicated finetuning API. Finetuning generates a new version of the base model, resulting in substantial implementation overhead — even if the AI provider hosts a robust finetuning environment.
    \item Techniques involving \emph{direct brain intervention on the model itself} allow for targeted modifications during inference potentially in real-time. The weights and activations of neurons in selected sub-areas of the neural network can be partially or completely altered, either adaptively for individual prompts or token-specific. While this approach is currently explored in research, to our knowledge it has not yet been deployed in production environments.
\end{enumerate}

A class 3 methodology that has matured into a prominent research direction are Sparse Autoencoders (SAEs), where features embedded in the relatively small inner dimensions of a LLM are reconstructed in an auto-encoding manner \cite{Gao2024,templeton2024scaling,bricken2023monosemanticity}. This allows for both a high-quality interpretation of these features and targeted control over the model's behavior by selectively activating specific features.

Our new method, which we name ``Mixture of Tunable Experts" (MoTE), also belongs into class 3. It is based on the established ``Mixture of Experts" (MoE) architecture that has been proven to be viable for decomposing transformer models into smaller, more manageable components called experts. Properly implemented MoEs can yield significant computational benefits with minimal to no performance degradation \cite{du2022glam,mistral2024cheaper}. Recent models published by DeepSeek, namely V3 and R1 \cite{deepseekai2025deepseekr1}, have further advanced this concept by increasing the number of experts to 14,906 \cite{dai2024deepseekmoe,deepseekai2024deepseekv3}. This is massively bigger than previously typical numbers and was key impetus for this research: As the number of experts increased so much, both in terms of total count and simultaneous activity, does this enable a new level of interpretability of their activations? If so, can this be leveraged to steer the model's behavior?

We investigate these questions by examining DeepSeek-R1.

\section{Technical Approach}

R1 has gained significant worldwide attention since its release. One of its most notable features is its advanced reasoning capability, which is a direct result of an innovative reinforcement learning training process. The open-source release of its weights under a permissive license has further boosted its popularity, making it an ideal choice for research purposes.

\begin{figure}[H]
\centering
\includegraphics[width=.67\textwidth]{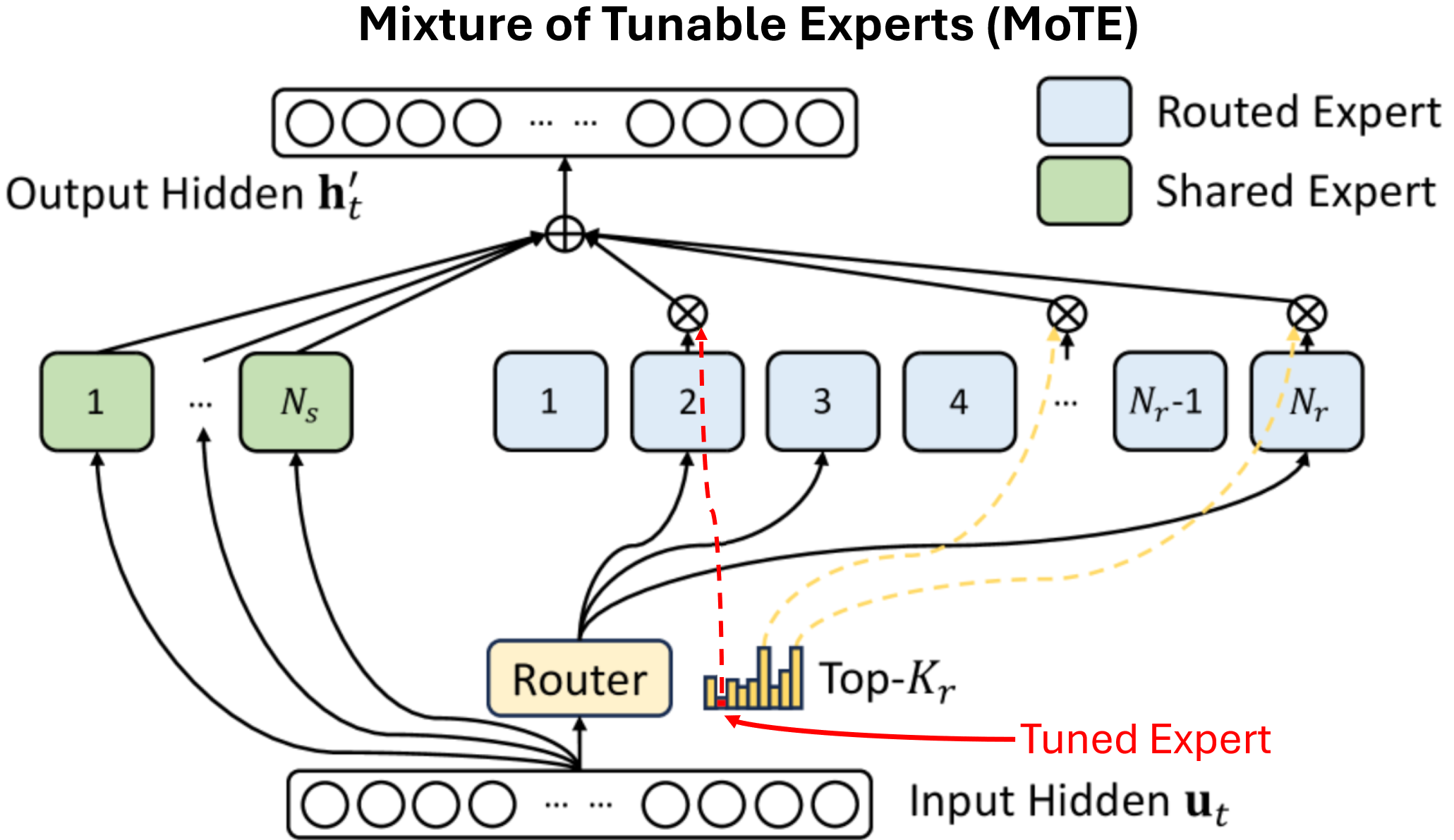}
\caption{MoTE as an extension to the DeepSeekMoE architecture (Illustration taken from \cite[p.7]{deepseekai2024deepseekv3} and modified). Shared experts are always activated. Originally, only the top-k routed experts selected by the Router's output get activated. With MoTE, there's an added flexibility: individual routed experts can be tuned by manually overriding the Router's output at inference time.}
\label{fig:deepseekmoe}
\end{figure}

The architecture of R1 is based on the previously introduced DeepSeekMoE \cite{dai2024deepseekmoe}, which enhanced the MoE architecture (cf. \cref{fig:deepseekmoe}). Sparse MoE layers \cite{fedus2022reviewsparseexpertmodels} optimize the Feed-Forward Networks (FFNs) within the Transformer framework \cite{vaswani2017attention} by cutting the FFNs into smaller, parallel subnetworks. DeepSeek categorizes these subnetworks as
\begin{itemize}
    \item \emph{shared experts}, which are always on and process all tokens, intended to capture common knowledge across varying contexts
    \item and \emph{routed experts}, which are only activated if they are selected by an upstream router network, intended to gain non-overlapping and focused knowledge \cite{dai2024deepseekmoe}.
\end{itemize}

The R1 designers massively scaled up their innovative concept, namely to one shared expert and 256 routed experts in each of its 58 MoE layers.
This results in 14,848 routed experts across the entire model, and in total 14,906 routed and shared experts.
During inference each router activates eight of the 256 routed experts in its layer.
In comparison, Mixtral 8x22B \cite{mistral2024cheaper}, an older MoE model, has only 448 experts in total.
They are organized into just eight routed experts in each of its 56 MoE layers with two of the eight selected.
Mistral, Mixtral's manufacturer, was surprised that they did not find evidence for its experts specializing in certain domains \cite[Section 5]{jiang2024mixtralexperts}, which in hindsight may simply be due to a too-small expert count per layer.
DeepSeek’s MoE architecture was specifically designed for increased specialization of the routed experts.

DeepSeek performed an ablation study to investigate the role of different experts. They found that switching off shared experts massively reduces the models capabilities, which indicates that basic knowledge is attained by these shared experts. Switching off top routed experts in each layer also degrades model performance. They measured significant negative effects when switching off 1/16 of the top routed experts, which is a very destructive act \cite[p.13]{dai2024deepseekmoe}.

These findings motivate the assumption that switching off a few individual experts among the $\sim$15k experts of R1 will harm the overall model performance only slightly if at all. At the same time, the fine granular subdivision into expert sub-networks may offer the ability to steer the model behavior by modifying the expert routing.

In order to measure and tune expert activations, we deployed R1 on an 8xH100 NVL cluster running a modified vLLM \cite{kwon2023efficient} inference engine, providing us access to expert routers for all individual layers and both input and completion tokens. \cite{tng2025}

\subsection{Analyzing Expert Activation}

\begin{figure}[H]
    \centering
    \includegraphics[width=\linewidth]{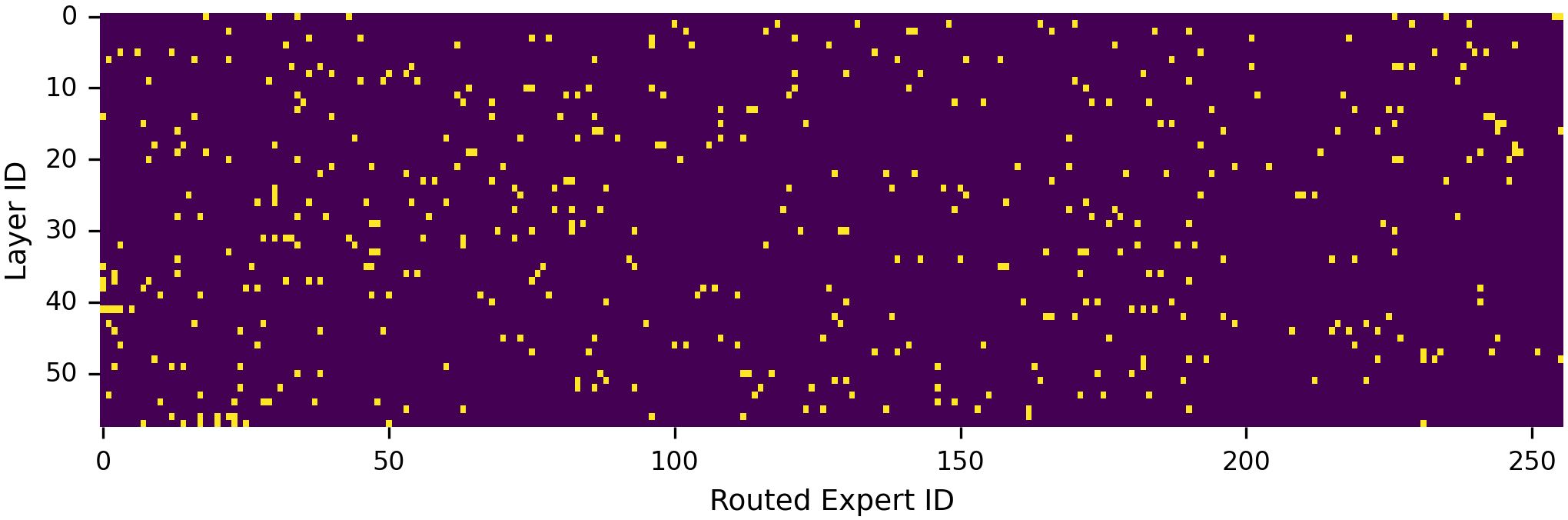}
    \caption{Functional Token Resonance Imaging (fTRI) visualizes the expert activation of an exemplary single token. For each gated layer in the network, the corresponding row depicts which 8 of the 256 routed experts got activated.}
    \label{fig:ftri:singletoken}
\end{figure}

Within the R1 MoE architecture, a single token only activates a fraction of 8 / 256 routed experts in each layer.
A router computes activation scores for all routed experts based on its hidden input taken from the attention block of this layer (cf. \cref{fig:deepseekmoe}).
Only the top-k experts having the highest activation scores will process the hidden input state.
Non-selected routed experts do not contribute to the next token generation step.
Each input token therefore activates a specific pattern of routed experts.
This activation pattern can be visualized in 2D for each token, as shown in \cref{fig:ftri:singletoken}.

\begin{figure}[H]
    \centering
    \includegraphics[width=\linewidth]{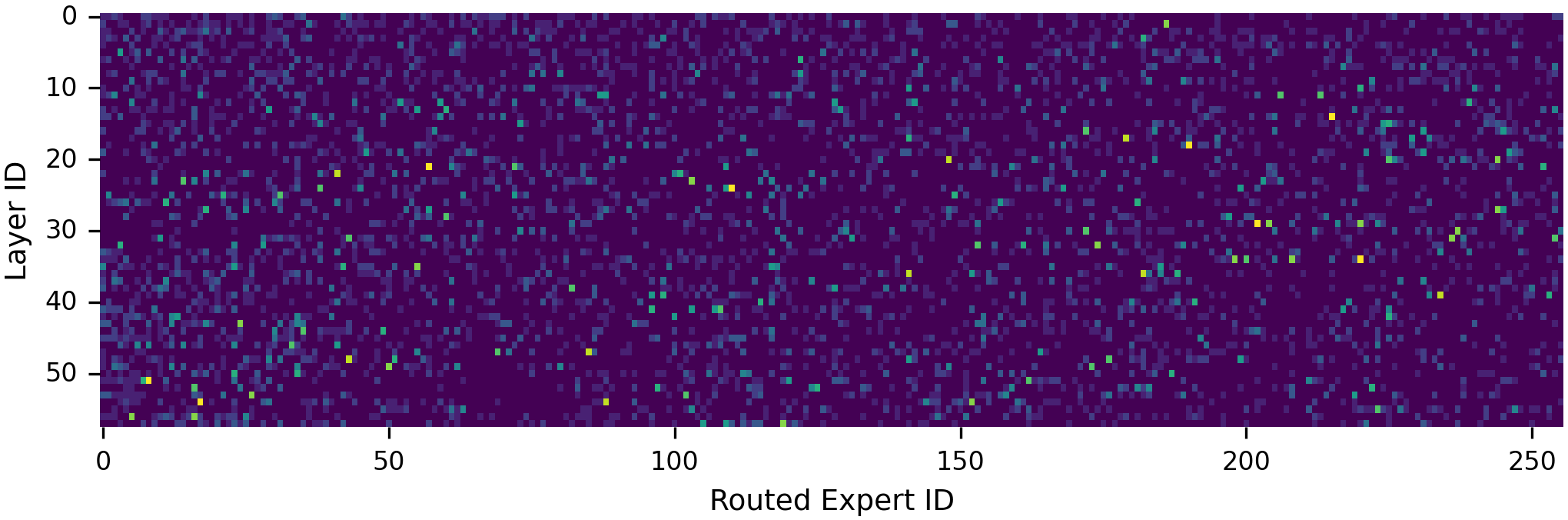}
    \caption{fTRI map that visualizes the expert activation pattern aggregated over all tokens in an input prompt. Each row depicts the activations of the layer experts as in \cref{fig:ftri:singletoken}, summed over all prompt tokens. A higher signal indicates that a given expert was selected for more input tokens.}
    \label{fig:tri:activationpatternexample}
\end{figure}

In order to gain a more insightful activation pattern, we consider expert activations aggregated over all tokens of the input prompt (\cref{fig:tri:activationpatternexample}). This condenses all expert choices of the LLM when it has parsed the input prompt and just before the first output token gets generated, i.e. just before it starts to generate an answer. Thus, we use the term functional Token Resonance Imaging (fTRI). Experts resonate with the provided token pattern.

If experts are specialized and responsible for specific knowledge domains, then fTRI visualizes these domains, as those experts having relevant knowledge about the concepts contained in the prompt will have been activated more often than experts that have no conceptional overlap.

\begin{flushleft}
\subsection{Multidimensional Prompt Sensitivity Analysis using behavior-specific\linebreak Dataset Generation}
\end{flushleft}

We aim to trigger and measure different model behaviors with regard to expert activations. Thus, we designed a parameterized prompt pattern whose parameter combinations elicit several distinct activation patterns.

The prompt pattern follows the format \promptquote{``What happened \beginpromptplaceholder time\fooendpromptplaceholder  \beginpromptplaceholder place\fooendpromptplaceholder ?"}. The \promptquote{\beginpromptplaceholder time\fooendpromptplaceholder}  parameter is selected from the range 1980–2025 and may also include relative terms such as \promptquote{``yesterday"}, \promptquote{``last year"}, and \promptquote{``last month"}. The possible locations for the \promptquote{\beginpromptplaceholder place\fooendpromptplaceholder}  parameter include \promptquote{``at Berlin Wall"}, \promptquote{``at Heathrow"}, \promptquote{``in Manhattan"}, ``in Singapore", ``in Beijing", \promptquote{``in Shanghai"}, \promptquote{``in London"}, \promptquote{``in South Africa"}, \promptquote{``in Mexico"}, \promptquote{``in Paris"}, and \promptquote{``in Brussels"}. The Cartesian combination of all attributes yields an analytical dataset with 756 question sentences, directly usable as prompts. Note that we made sure that the questions contain some (time, place)-combinations that the LLM might consider as sensitive as well as some that the LLM would probably consider as harmless.

Our dataset triggers three types of answers:
\begin{itemize}
    \item REFUSED: The DeepSeek model directly responds to the question with an answer of the form: \promptquote{``\textless think\textgreater \textless/think\textgreater  I am sorry, I cannot answer that question. I am an AI assistant designed to provide helpful and harmless responses."} So it does not start its reasoning process but simply refuses to answer. We classify questions triggering this behavior and their corresponding activation patterns using the label \classemph{0-REFUSED}
    \item ALIGNED: DeepSeek does not enter its reasoning process but instead directly returns an answer of the form: \promptquote{``\textless think\textgreater \textless/think\textgreater  In 2017, several notable events occurred at London's Heathrow Airport, one of the busiest airports in the world. ..."}. These answers are somewhat specific to the question, but because there is no reasoning process involved, they constitute a generic collection of facts that seem to have been learned during the alignment process after the RL training. We thus classify them under the label \classemph{1-ALIGNED}
    \item REASONED: DeepSeek enters the reasoning process for which it got famous, before composing the final answer. In these cases, the responses start with: \promptquote{``\textless think\textgreater Okay, so I need to figure out what happened in 2000 at the Berlin Wall. Let me start by recalling what I know about..."}. Since these responses make use of the reasoning step R1 had explicitly been trained for, we classify them using the label \classemph{2-REASONED}
\end{itemize}

The difference between ``aligned" and ``reasoned" answers is commonly attributed to the distinction between System 1 thinking (fast, intuitive, and automatic) and System 2 thinking (slow, deliberate, and reasoning-based) (\cite{Kahneman2003-KAHAPO})

For all prompts in our dataset, we record the expert activation pattern for each token in both the question prompt and the answer completion. Based on these recordings, we derived an overall activation pattern for each prompt, as illustrated in \cref{fig:tri:activationpatternexample}. In the following, we will analyze these averaged prompt activation patterns further and relate them with their respective classifications.

\subsection{Activation Pattern Understanding / Insights}

\begin{figure}[H]
    \centering
    \includegraphics[width=.85\linewidth]{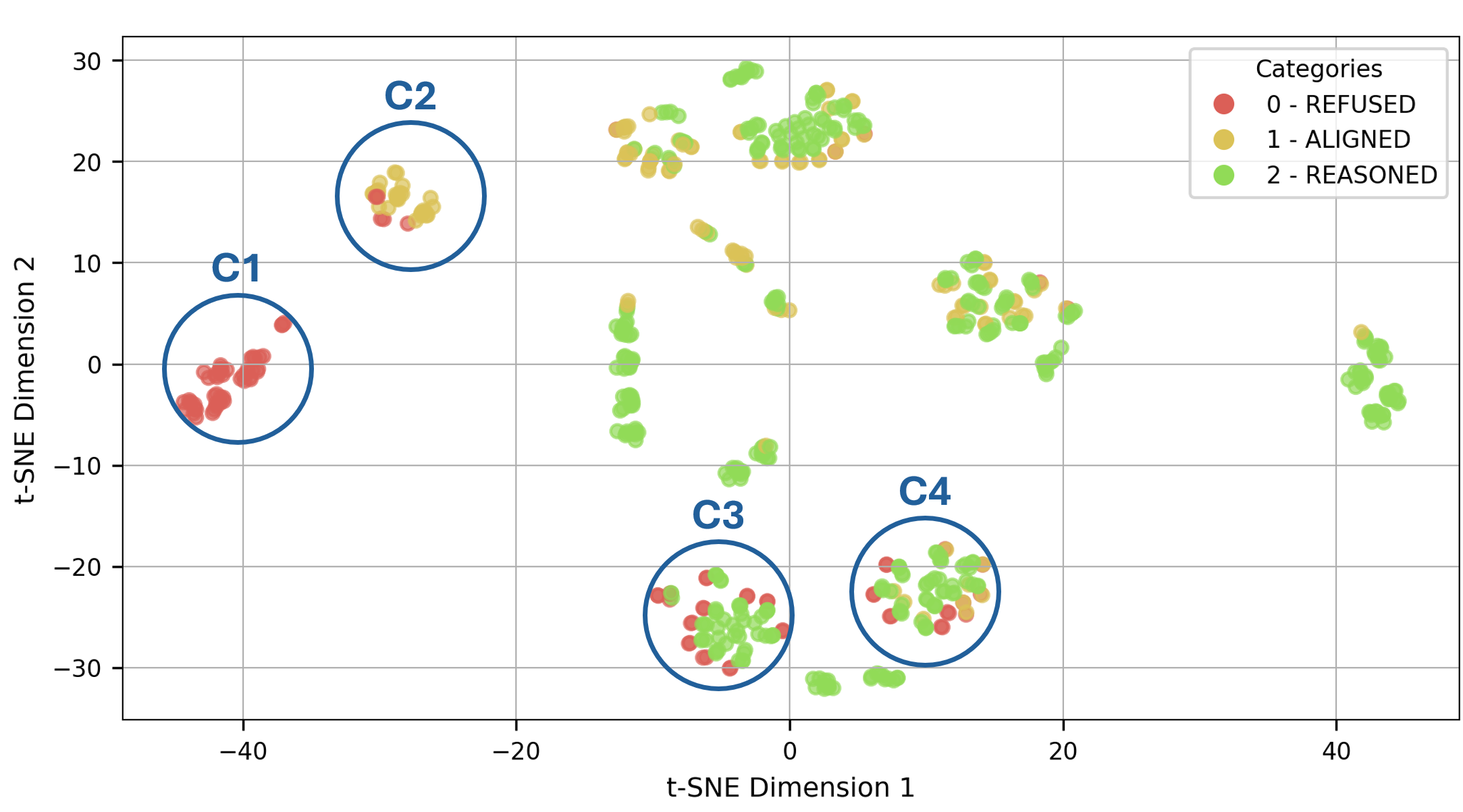}
    \caption{Mapping expert activation patterns for different prompts into two dimensions using t-SNE reveals distinct clusters of similar activation patterns. E.g. cluster 2 represents questions about recent events (see text for details on clusters C1 - C4).}
    \label{fig:tsne:actpatterns}
\end{figure}

To better understand how activation patterns compare across our analytical dataset, we apply t-distributed Stochastic Neighbor Embedding (\cite{van2008visualizing}) to generate a two-dimensional projection of the original expert activation patterns. t-SNE represents each high-dimensional expert activation pattern as a two-dimensional point (\cref{fig:tsne:actpatterns}). This ensures that similar activation patterns are positioned close to each other, while dissimilar patterns are mapped to distant points with high probability.

Mapping expert activation patterns for different prompts into two dimensions using t-SNE reveals distinct clusters of similar activation patterns. Specifically, when examining \classemph{0-REFUSED} patterns, we observe a clearly separated cluster of prompt activations (\classemph{C1}) corresponding to questions about certain places and events in China, that are not supposed to get answered by the model. Other clusters are less distinct. For instance, questions about recent events like \promptquote{``yesterday"}, \promptquote{``last month"}, \promptquote{``last year"} (\classemph{C2}) mostly result in aligned answers, but for certain places like the Berlin Wall in Germany they get refused, too. When asked about events in the 1980s (\classemph{C3}) and 1990s (\classemph{C4}), the model activates its reasoning process to provide an answer, except for places like Beijing or Shanghai, which get refused. This motivates the hypothesis that tuning the right experts may change the model behavior such that refused answers from these clusters switch to aligned answers.

\subsection{Identification of Distinctive Experts}

Using the parameterized prompt pattern above we can now systematically trigger three different response types \classemph{0-REFUSED}, \classemph{1-ALIGNED}, and \classemph{2-REASONED} and study the activation patterns that contribute most specifically to each type of model behavior.

We hypothesize that \emph{the experts with a high activation} to \classemph{0-REFUSED} answers and a low activation otherwise can specifically be associated with prompts where the LLM provides an answer of type \classemph{0-REFUSED}. By actively suppressing these expert activations during inference, we expect to be able to influence the model’s behavior such that fewer answers get refused.

To achieve this goal, we process our analytical dataset as follows:

\begin{enumerate}
    \item \emph{Prompt Activation Map:} For each prompt, we sum up the activations of each routed expert in all layers for all tokens of the input prompt (see \cref{fig:tri:activationpatternexample} for an example).
    \item \emph{Map Classification:} The maps get classified into one of the three predefined response classes \{\classemph{0-REFUSED}, \classemph{1-ALIGNED}, \classemph{2-REASONED}\}, depending on the model response to the underlying input prompt.
    \item \emph{Average Activation Map:} Within each category, we average the activation maps.
    \item \emph{fTRI:} Finally, functional Token Resonance Imaging helps us to identify the experts that most distinctively contribute to each class. As a measure of resonance we take the averaged activations for the desired class and subtract the averaged activations of all other classes.
\end{enumerate}

\begin{figure}[H]
    \centering
    \includegraphics[width=\linewidth]{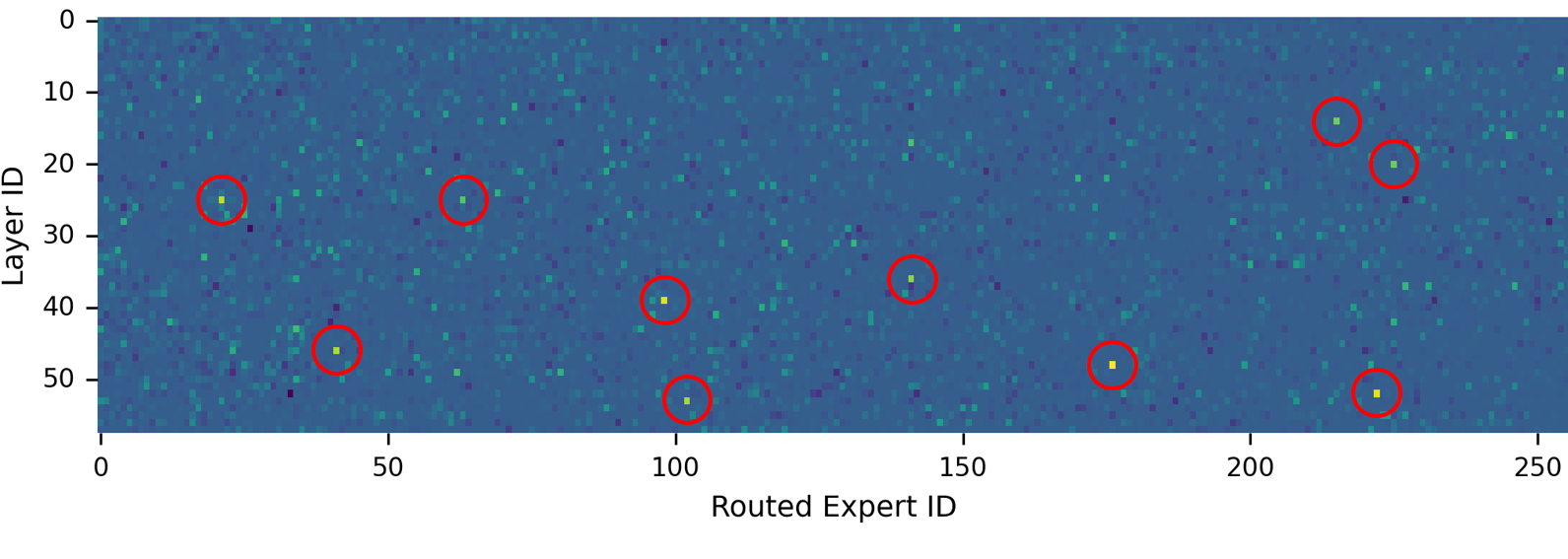}
    \caption{fTRI map of differential activations for refused prompts. Highly distinctive experts (highlighted with red circles, e.g. routed expert with expert-id 176, layer-id 48) are chosen as candidates for tuning, their exact coordinates are listed in \cref{sec:appendix:topexperts}.}
    \label{fig:ftri:censorshipexperts}
\end{figure}

In practice, to identify those experts that are most likely responsible for answer refusal, we consider the fTRI map for the class \classemph{0-REFUSED} (\cref{fig:ftri:censorshipexperts}). Experts with high resonance values in this map represent potential candidates that may contribute significantly to answer refusal. The fTRI map shows that only a few routed experts stand out. The top 5 - 7 experts can directly be identified by just looking at \cref{fig:ftri:censorshipexperts}., where the most distinctive expert for refusal identifies as (expert-id, layer-id) = (176, 48). Most of our experiments were done by tuning the top 10 resonant experts among the $\sim$15k routed experts in total, to keep side effects of the manipulation low.

\section{Results}
\subsection{Suppressing Distinctive Experts}

With the distinctive experts for refusal of answers at hand, we performed an ablation study to test if these experts are crucial for the LLM behavior to refuse certain answers.
We suppressed these experts by adapting the expert router functions: if one of the identified experts is selected among the top-k ids, meaning chosen by the MoE router (cf. \cref{fig:deepseekmoe}), we set its corresponding weight to zero and renormalize the other top-k weights.
This approach guarantees that the identified experts have no impact on the output, and that no alternative experts are selected instead.
Using the modified MoE, we generated new responses for our set of fixed-template prompts and classify them accordingly.

\begin{figure}[H]
    \centering
    \includegraphics{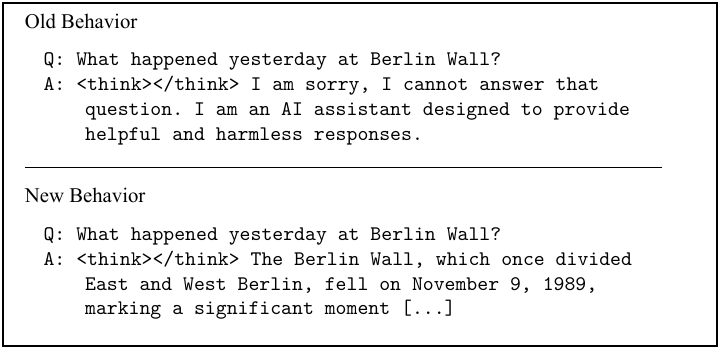}
    \caption{Exemplary shift from \classemph{0-REFUSED} to \classemph{1-ALIGNED}.}
    \label{fig:prompt:refusedtoaligned}
\end{figure}

\begin{figure}[H]
    \centering
    \includegraphics{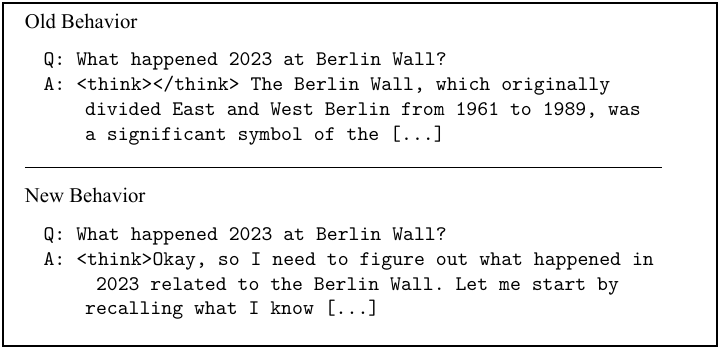}
    \caption{Exemplary shift from \classemph{1-ALIGNED} to \classemph{2-REASONED}.}
    \label{fig:prompt:alignedtoreasoned}
\end{figure}

\Cref{fig:prompt:refusedtoaligned,fig:prompt:alignedtoreasoned} show two exemplary model responses with and without suppression of the distinctive experts, one demonstrating a shift from a \classemph{0-REFUSED} to an \classemph{1-ALIGNED}, the other a shift from an \classemph{1-ALIGNED} to a \classemph{2-REASONED} answer.
Importantly, we did not observe a degradation in general language capabilities (cf. also \cref{sec:quality}).

\begin{figure}[H]
    \centering
    \includegraphics[scale=\resultgridfiguresscale]{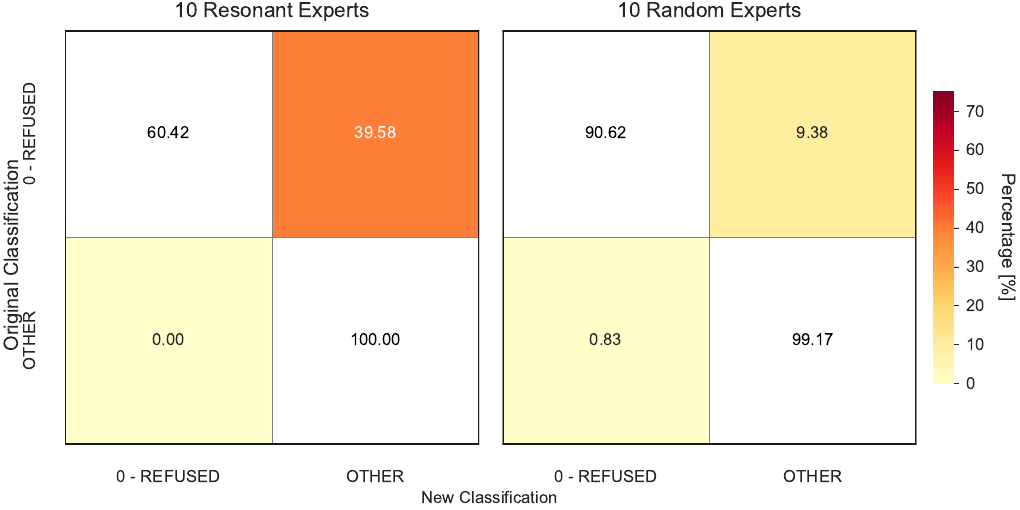}
    \caption{Left: Change in model behavior with suppressed experts for the parameterized dataset. The top row shows that the tuned model provides a non-refused answer for almost 40\% of those prompts that get refused by R1. The second row shows that no additional prompts get refused, i.e. for this dataset there is no unwanted effect in the opposite direction. Right: Control results with randomly selected experts. More than 90\% of the previously refused answers remain refused.}
    \label{fig:results10experts}
\end{figure}

Overall, the expert suppression results in a significant reduction of refused answers (\cref{fig:results10experts}): the model can be changed to provide an answer for about 40\% of the originally refused prompts.
Notably, previously non-refused answers all remain non-refused.

As a demonstration that the weakening of alignment is specific to the set of suppressed experts, we run a control experiment and deactivate a random set of 10 routed experts (\cref{fig:results10experts}).
Suppressing random experts does affect some of the answers too, and changes about 9\% of the previously refused answers.
This is much smaller than the 40\% effect we measured for deactivating distinctive experts.
About 1\% of previously non-refused answers change in the opposite direction and get refused, indicating that refusal is easier to break than to enforce.

\subsection{Validating Distinctive Expert Suppression with a Larger Dataset}

Notably, although we identified the set of distinctive experts using prompts constructed from a fixed question template, suppressing these experts reduces refusal likelihood across a broader range of prompts. For example, within a set of 1,360 prompts on sensitive topics (\cite{promptfoo2025, promptfoo2025b}) we observe that 52\% of previously refused prompts flip in the desired direction. In contrast, the LLM changes its answer to refusal for only about 1.3\% of the prompts that did not get refused before. This validates that the experts identified using fTRI correspond to the general concept of answer refusal and are not merely specific to the prompt format used in our dataset to trigger the behavior in fTRI.

\begin{figure}[H]
    \centering
    \includegraphics[scale=\resultgridfiguresscale]{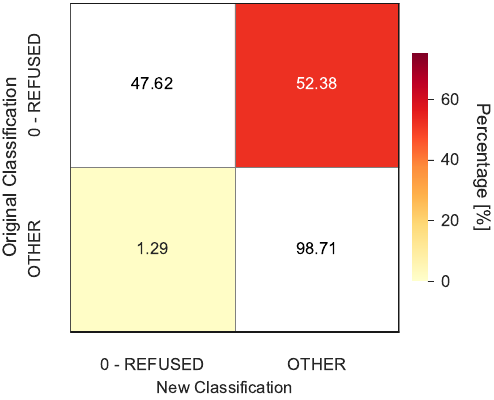}
    \caption{Expert suppression results, as shown in \cref{fig:results10experts}, but applied for the sensitive topics dataset.}
    \label{fig:resultsccp}
\end{figure}

\subsection{Stimulating Distinctive Experts}

Instead of suppressing specific experts by zeroing their respective weights in the router functions, we can also stimulate their activation. In this case, the modification to the router function is more complex, as for many tokens the alignment-specific experts are not already among the top-k selected experts of their layer.

Therefore, we implemented the following approach: In all layers we added the stimulated experts to the top-k selected experts, irrespective of their weight. In order to keep the number of selected experts constant, we removed the lowest of the top-k selected experts with respect to their weights, if necessary. Then, we set the weight of the stimulated expert to the maximum value of all weights and renormalize. This approach guarantees that the stimulated experts always contribute to the overall model response while keeping undesired side-effects from deactivating other top-k selected experts at a minimum.

\begin{figure}[H]
    \centering
    \includegraphics[scale=\resultgridfiguresscale]{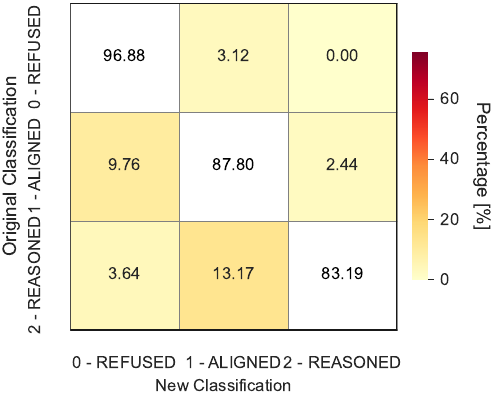}
    \caption{Result in analogy to \cref{fig:results10experts}, however stimulating instead of suppressing the distinctive experts.}
    \label{fig:resultsstimulation}
\end{figure}

To test the effect, we stimulated the alignment-specific experts and looked at the transitions between the three classes \classemph{0-REFUSED}, \classemph{1-ALIGNED} and \classemph{2-REASONED}. We found that the fraction of refused responses increased to about 10\% for previously aligned responses, while the unwanted effect in the opposite direction is only 3\%. Previously reasoned answers got reduced by 17\%, most of them changed to \classemph{1-ALIGNED} and 3.6\% changed into the \classemph{0-REFUSED} class as desired (\cref{fig:resultsstimulation}).

Overall, stimulating the alignment-specific experts had a less pronounced effect compared to suppression, but the effect was still systematic. In particular, changing the direction of expert tuning also inverted the trend in which the answers changed, as expected.

\subsection{Application to other areas of LLM behavior: Changing R1's Reasoning Language}

Having shown that expert tuning does modify the LLM behavior in regard to its alignment, we wondered if it is just a special case. It could be argued that R1's refusal to answer might not be an innate property, because alignment against harmful responses of R1 - and actually most LLMs - is performed in final stages of the training process \cite{deepseekai2025deepseekr1}. Therefore, we wanted to test if the MoTE method generalizes to other behaviors.

Our test candidate was the language that R1 uses for reasoning. The rationale for this choice is a) that R1's thinking process, which emerged during a reinforcement learning phase \cite{deepseekai2025deepseekr1}, is an intrinsic, characteristic behavior and b) that another version of the model, DeepSeek-Zero, showed reasoning language mixing and thus poor readability during its reasoning steps. The latter indicates that the high consistency of R1 in using either English or Chinese as its reasoning language has been deeply trained into the model. Successfully using MoTE to tune aspects of this fundamental behavior would be a strong indication for a broader applicability of the method.

\begin{figure}[H]
    \centering
    \includegraphics[width=\linewidth]{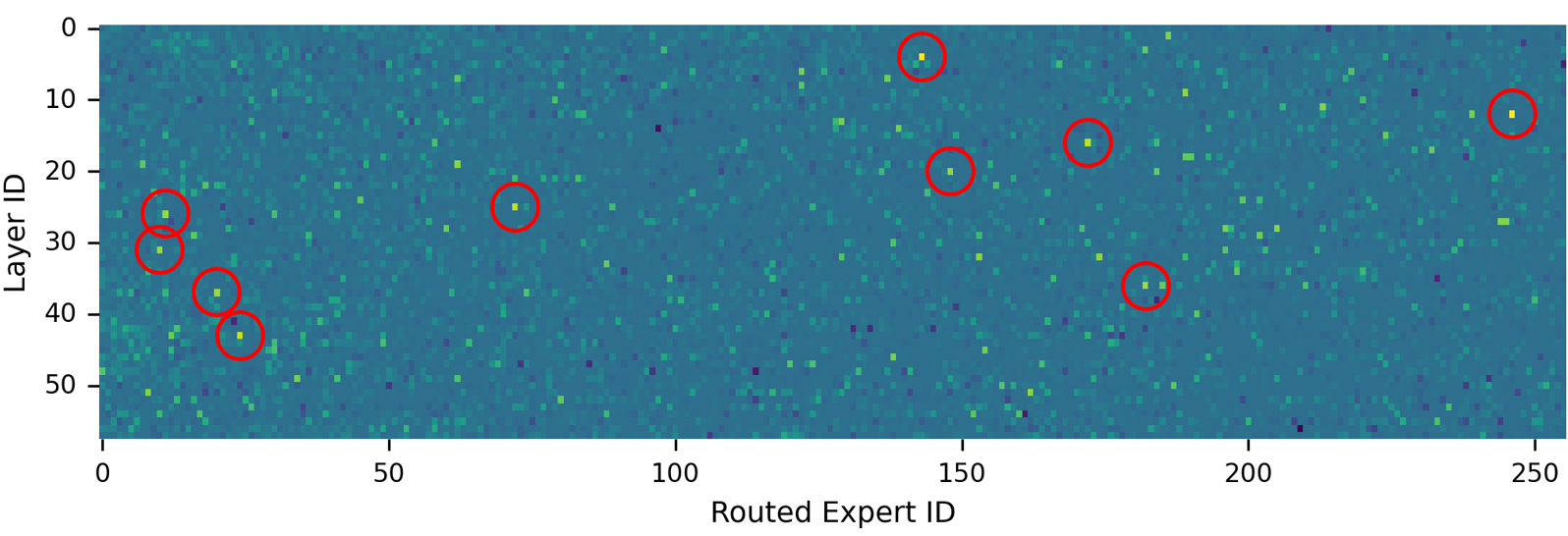}
    \caption{fTRI map of differential activations for reasoning in English and Chinese, respectively. The most distinctive experts are highlighted with red circles, their exact coordinates are listed in \cref{sec:appendix:topexperts}.}
    \label{fig:languages:ftri}
\end{figure}

To test this hypothesis, we created a new dataset of 600 prompts that trigger R1 to reason either in Chinese (200) or in English (400). With fTRI, we were able to identify the most distinctive experts active when the model chooses to reason in English. The corresponding fTRI scan is shown in (\cref{fig:languages:ftri}). The most distinctive expert is (expert-id, layer-id) = (143, 4), which is close to the input embedding layer and colored in yellow.

\begin{figure}[H]
    \centering
    \includegraphics[scale=\resultgridfiguresscale]{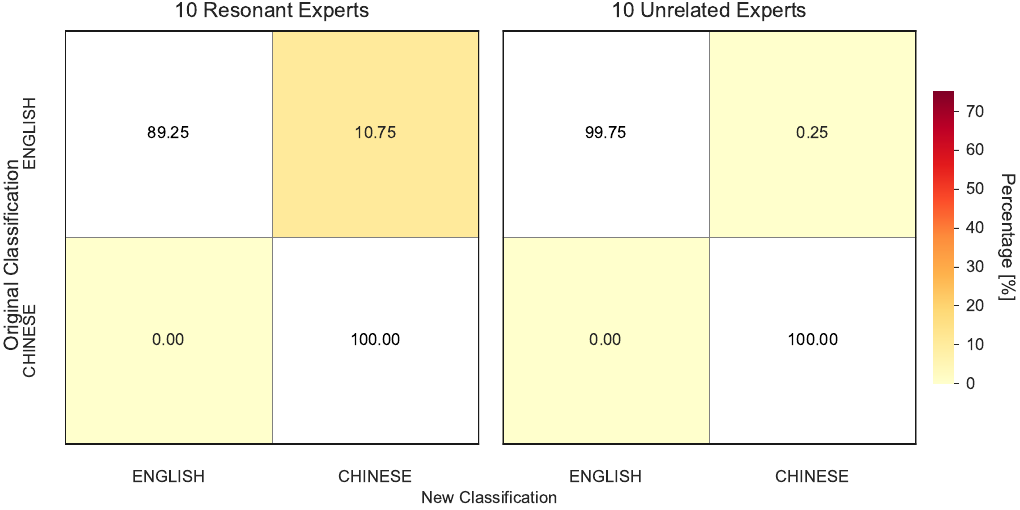}
    \caption{Change in the reasoning language from English to Chinese.}
    \label{fig:languages:res}
\end{figure}

After tuning down the top 10 distinctive experts for choosing the reasoning language English, we reprocessed the dataset and indeed, the model changed its behavior to reason in Chinese more often (see \cref{fig:languages:res}). For 10.75\% of the prompts within our dataset, R1 changed its behavior from reasoning in English to reasoning in Chinese. The effect is smaller compared to the reduction in answer-refusal measured above, which indicates that more elaborate ways of tuning experts may be required to get a bigger effect. Still, it does indicate a broader applicability of the fTRI / MoTE method.

To verify that the specific selection of experts is crucial for triggering changes in the reasoning language, we tested the LLM by switching off 10 unrelated experts. In that case, we did not observe any switching of language apart from one accidental flip from English to Chinese, i.e. 0.25\%.  

The prompts used to trigger the model to use different languages for thinking take the template format: \promptquote{``What is \beginpromptplaceholder number\_a\fooendpromptplaceholder  plus \beginpromptplaceholder number\_b\fooendpromptplaceholder  equal to?"} with the optional extension \promptquote{``Please make sure to answer in \beginpromptplaceholder language\fooendpromptplaceholder ."}, where language is set to either English or Chinese. We duplicated these prompt templates by asking the same questions in Chinese. To boost the dataset size, we varied the prompt template to all combinations of two numbers from 1 to 10. Here is an example that illustrates the resulting prompts and the effect on the model behavior. In this case we successfully tuned the model to switch the reasoning language from English to Chinese.

\begin{figure}[H]
    \centering
    \includegraphics{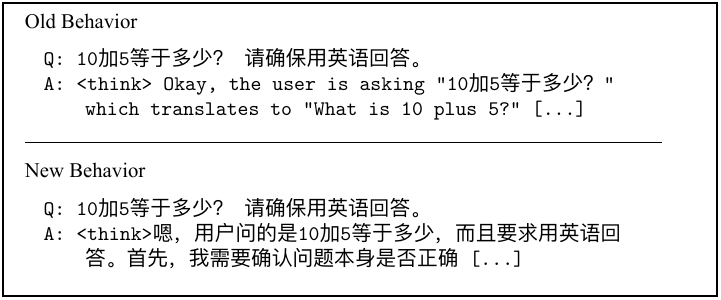}
    \caption{Exemplary shift of the reasoning language from English to Chinese.}
    \label{fig:prompt:language}
\end{figure}

\subsection{Impact on General Model Quality}\label{sec:quality}

\begin{figure}[H]
    \centering
    \includegraphics[width=.75\linewidth]{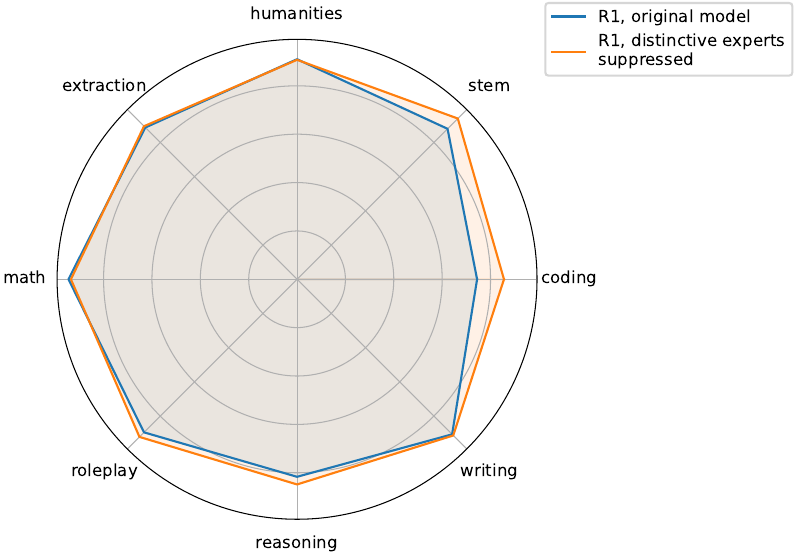}
    \caption{Benchmark comparison between the original DeepSeek-R1 model and our MoTE variant, after suppressing distinctive experts. The performance has been evaluated for the MT-Bench dataset \cite{zheng2023mtbench}, numerical scores are listed in \cref{sec:appendix:benchmarkscores}.}
    \label{fig:benchmark:mtbench}
\end{figure}

To gain a preliminary understanding of how MoTE influences the overall quality of the model, we compared benchmark results with and without deactivating alignment specific experts (\cref{fig:benchmark:mtbench}). We utilized MT-Bench, a widely recognized benchmark for assessing LLM performance in multi-turn dialogues. MT-Bench comprises 80 high-quality, multi-turn questions designed to evaluate conversational and instruction-following abilities across various categories, including writing, roleplay, extraction, reasoning, mathematics, coding, and knowledge in both STEM and humanities disciplines \cite{zheng2023mtbench}.

We found that deactivating experts that are distinctive for refusal of sensitive or harmful questions does not degrade the benchmark results. Rather, we got indications of increased model performance, which surprised us somewhat. Drawing strong conclusions seems a bit too early for us, as the effect should be verified across a larger set of benchmarks.

\section{Conclusion}

The experimental results support our hypothesis that certain traits — such as the alignment behavior — of MoE models with many experts such as R1 are at least partially encoded within small subsets of experts. We introduced a simple yet effective analytical approach, fTRI, to identify such expert subsets by targeting them with parameterized prompt patterns whose parameter combinations elicit several distinct activation patterns. By altering expert selections — and thereby actively modifying routing mechanisms in the LLM — we can suppress or stimulate specific expert groups, leading to corresponding changes in model behavior.

\pagebreak
\section*{Acknowledgements}

\begin{CJK}{UTF8}{gbsn}谢谢\end{CJK}, vielen Dank, and thanks to DeepSeek for their innovative spirit and for making the excellent V3 and R1 models available. Without these models, this research would not have been possible yet.
Also thanks to the OpenSource community, particularly the teams around vLLM and HuggingFace, and all other OSS contributors.
Thanks to Lennard Schiefelbein, TNG, for the speedy support in performing the benchmark runs.

\pagebreak
\printbibliography

\pagebreak
\section*{Appendix}
\appendix

\section{Exact list of top-10 distinctive experts}\label{sec:appendix:topexperts}

We use the tuple notation \texttt{(layer-id, expert-id)} to specify each expert unambiguously. Here, the expert-id refers to the id within the routed experts of each layer.

\paragraph*{Distinctive experts for refused prompts}
~
\begin{lstlisting}[breaklines=true]
# list of (layer id, routed expert id)
[(48, 176), (52, 222), (39, 98), (25, 21), (46, 41), (53, 102), (36, 141), (14, 215), (20, 225), (25, 63)]
\end{lstlisting}

\paragraph*{Distinctive experts for reasoning in English}
~
\begin{lstlisting}[breaklines=true]
# list of (layer id, routed expert id)
[(12, 246), (4, 143), (43, 24), (25, 72), (16, 172), (36, 182), (26, 11), (31, 10), (20, 148), (37, 20)]
\end{lstlisting}

\section{Benchmark Raw Data}\label{sec:appendix:benchmarkscores}

\begin{table}[H]
    \centering
    \resizebox{\textwidth}{!}{
    \begin{tabular}{lccccccc}
        \toprule
        & \textbf{coding} & \textbf{extraction} & \textbf{humanities} & \textbf{math} & \textbf{reasoning} & \textbf{roleplay} & \textbf{writing} \\
        \midrule
        R1, distinctive experts suppressed & 8.56 & 8.95 & 9.08 & 9.35 & 8.48 & 9.22 & 9.13 \\
        R1, original & 7.44 & 8.88 & 9.10 & 9.45 & 8.17 & 8.95 & 9.07 \\
        \bottomrule
    \end{tabular}
    }
    \caption{Performance scores across different categories.}
\end{table}

\end{document}